\documentclass[10pt,twocolumn,letterpaper]{article}

\usepackage{pdfpages} 

\usepackage{cvpr}
\usepackage{times}
\usepackage{epsfig}
\usepackage{graphicx}

\usepackage[T1]{fontenc}
\usepackage[latin9]{inputenc}
\usepackage{array}
\usepackage{verbatim}
\usepackage{rotating}
\usepackage{url}
\usepackage{multirow}
\usepackage{amsmath}
\usepackage{amssymb}
\usepackage{graphicx}

\usepackage{times}
\usepackage{helvet}
\usepackage{courier}
\usepackage{algorithm}
\usepackage{algpseudocode}

\newcommand{\action}{w}
\newcommand{\actiont}{\action^{(t)}}
\newcommand{\rw}{r_{\actiont}}
\newcommand{\btheta}{\boldsymbol\theta}
\newcommand{\func}{f_{\boldsymbol\theta}(\actiont)}

\newcommand{\bactions}{\mathbf{\mathcal{W}}}
\newcommand{\brewards}{\mathbf{\mathcal{R}}}

\newcommand{\mean}{\hat{\mu}(\actiont)}
\newcommand{\var}{\hat{\sigma}^2(\actiont)}
\newcommand{\std}{\hat{\sigma}(\actiont)}

\newcommand{\regrett}{\rho_t}
\newcommand{\Regrett}{\rho'_T}
\renewcommand{\vec}[1]{#1}
\DeclareMathOperator*{\argmax}{\arg\!\max}

\newcommand{\Tdialogue}{T_{\text{dialogue}}}
\newcommand{\hist}{D}
\newcommand{\histt}{\hist_t}

\newcommand{\Y}{O}
\newcommand{\y}{o}
\newcommand{\ystar}{o^*}

\def\fl[#1\]{\begin{align}#1\end{align}}
\def\[#1\]{\begin{align*}#1\end{align*}}
\def\*[#1\]{\begin{align*}#1\end{align*}}

\usepackage{color}

\usepackage{bm}

\newcommand{\stopdialoguetoken}{{{<}stop{>}}}

\usepackage[pagebackref=true,breaklinks=true,letterpaper=true,colorlinks,bookmarks=false]{hyperref}

\cvprfinalcopy 

\def\cvprPaperID{1771} 

\ifcvprfinal\fi
\begin{document}

\title{What's to know? \\ Uncertainty as a Guide to Asking Goal-oriented Questions}

\author{Ehsan Abbasnejad, Qi Wu, Javen Shi, Anton van den Hengel \\
\texttt{\small{}\{ehsan.abbasnejad,qi.wu01,javen.shi,anton.vandenhengel\}@adelaide.edu.au} \\ 
 Australian Institute of Machine Learning \&
 The University of Adelaide, Australia
}

\maketitle

\begin{abstract}
One of the core challenges in Visual Dialogue problems is asking the question that will provide the most useful information towards achieving the required objective.  
Encouraging an agent to ask the right questions is difficult because we don't know a-priori what information the agent will need to achieve its task, and we don't have an explicit model of what it knows already.
We propose a solution to this problem based on a Bayesian model of the uncertainty in the implicit model maintained by the visual dialogue agent, and in the function used to select an appropriate output.  By selecting the question that minimises the predicted regret with respect to this implicit model the agent actively reduces ambiguity.
%
The Bayesian model of uncertainty also enables a principled method for identifying when enough information has been acquired, and an action should be selected.
We evaluate our approach on two goal-oriented dialogue datasets, one for visual-based collaboration task and the other for a negotiation-based task. Our uncertainty-aware information-seeking model outperforms its counterparts in these two challenging problems.
\end{abstract}


\vspace{-5pt}
\section{Introduction }

One of the fundamental problems in any challenge that requires actively seeking the information required to carry out a task is that of identifying the information that will best enable the agent to achieve its objective.  Identifying the information needed, and how to get it, is inherently complex, not least because the space of all possibly useful information is so large.  We propose a solution to this problem here that is applicable to reinforcement learning in general, and that we demonstrate on the challenging problem of goal-oriented visual dialogue.

Goal-oriented visual dialogue requires the participants to engage in a natural language conversation towards a specified objective.  The objectives of the two participants might be collaborative, such as communicating the identity of a specific object in an image~\cite{guesswhat_game}, or they may be adversarial~(see Sec.~\ref{sec:DealOrNoDeal}).  The challenge thus represents a close analogue to a visual version of the Turing test, and one of the key problems in computer vision that is closest to requiring Artificial Intelligence.  

The primary technological challenge in goal-oriented visual dialogue is to devise natural language interactions that are directed towards achieving the required objective.  This is in contrast to the more traditional approach that aims only to keep the other participant talking for as long as possible~\cite{li2016deep,serban2016building,neural_conv}.  Note that the performance criteria in these two approaches are opposite, as in goal-directed visual dialogue success is indicated by achieving the shortest possible conversation.



\begin{figure}
	\centering\includegraphics[width=1.05\columnwidth]{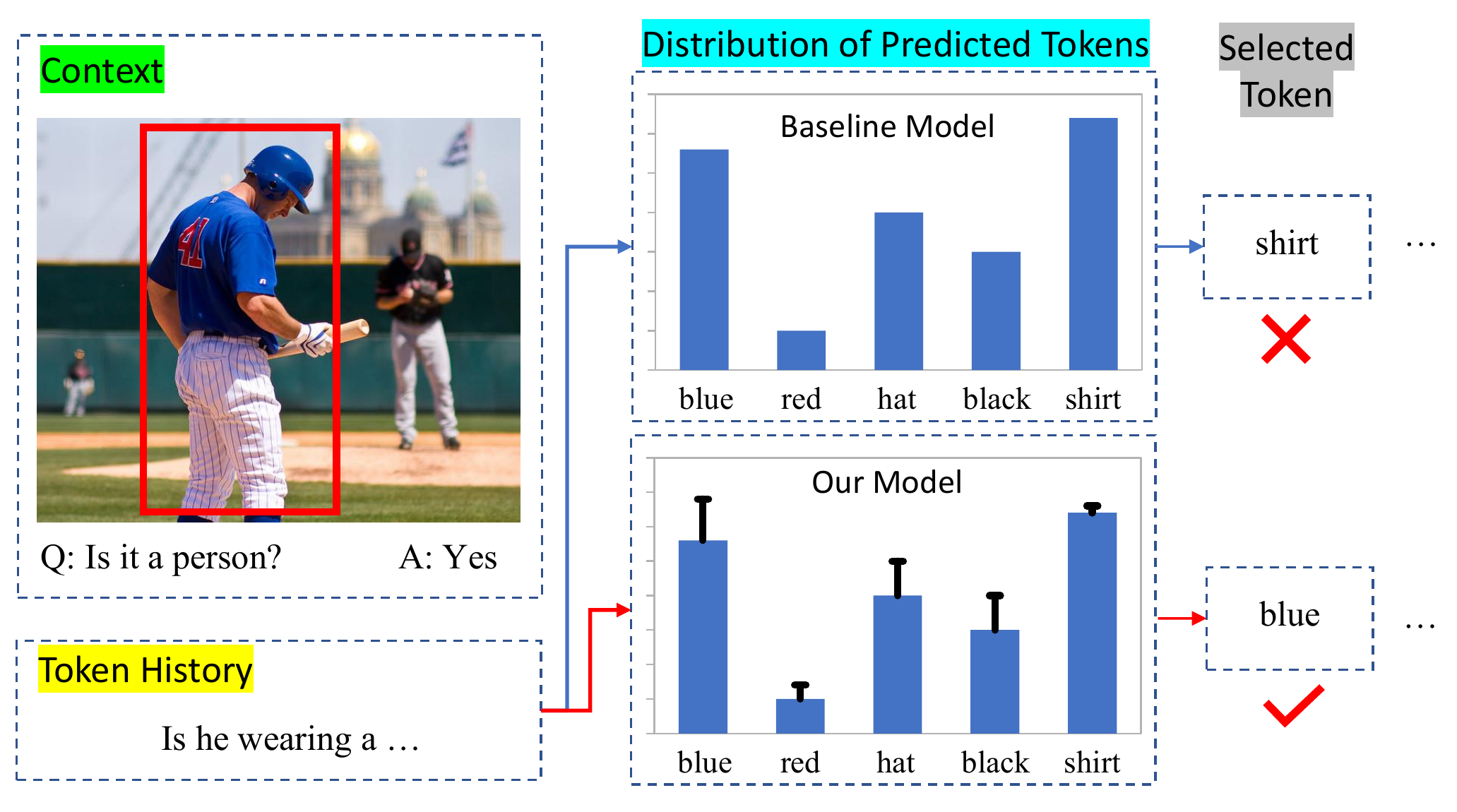}
	\caption{
	In GuessWhat~\cite{guesswhat_game} one player knows the correct object (here shown in a red box), and the other must ask questions to identify it.  Traditionally an agent would generate questions by sequentially selecting words with the highest conditional probability, even though knowing the answer might be uninformative (in this case `shirt' in the Baseline histogram). Our solution, however, selects `blue', which corresponds to the highest sum of the probability and standard deviation (likely to be the most `informative').
	}
	\vspace{-15pt}
	\label{fig:intro}
\end{figure}
Inspired by the success of deep learning in both computer vision and
natural language processing (NLP), most recent goal-oriented dialogue
studies rely on sequence-to-sequence (seq2seq) deep learning models
\cite{seq2seq,neural_conv}.
Obtaining the large datasets this approach requires is challenging, however.
As a partial solution, a combination of seq2seq and  deep reinforcement learning \cite{rl_intro} 
are are commonly used to train a model (\ie agent) with unlimited self-generated 
data in a self-play environment.
Even if this was achieved ideally, however, it is unlikely that it would lead to an agent capable of carrying out the complex reasoning needed to devise the next interaction that will recover exactly the information required to achieve an as yet unspecified objective.

Ideally, rather than learning to generate questions solely by reinforcement learning, the method should calculate the question that, when answered, 
will provide the most useful information for achieving the agent's objective.
The direct approach would require enumerating everything the agent might ever need to know, and the value of each such piece of information 
towards achieving its as yet unspecified objective.
This would allow the identification of the missing piece of information that is most critical to achieving the agent's objective, and the formulation of a corresponding question.

This direct approach is infeasible because the agent has the capacity to store all of the information it might need to hold about the task, the intention of its counterpart, the image, and so on.  Additionally, in the current state of the art approaches, this information is stored implicitly in the weights of a neural network.  Defining the scope of such an information store is impossible, which makes measuring its information content infeasible.  Explicitly relating the information stored to the agent's objective is similarly infeasible. This makes it impossible to directly identify the question that will provide the most useful information towards achieving the agent's objective.

Visual dialogue models trained using reinforcement learning already learn to estimate the value of a particular question as a step towards achieving their objective.  This is represented in the model's value function.
All that is required is a method for identifying the gaps in the model's internal information. We could then combine these information gaps with the learned policy to identify the most useful question.  

Given that the models in question represent their internal information implicitly, a good approximation of the model's information gaps is available in the uncertainty of its internal state.
  By propagating the model's internal uncertainty through the question generation process we can thus identify questions that best reflect the model's ambiguity in achieving its objective.  This is as compared to the current process that selects the question the model is most certain about (see Fig.~\ref{fig:intro}).

\begin{figure*}[t]
\centering

\vspace{-3mm}
\includegraphics[width=\textwidth]{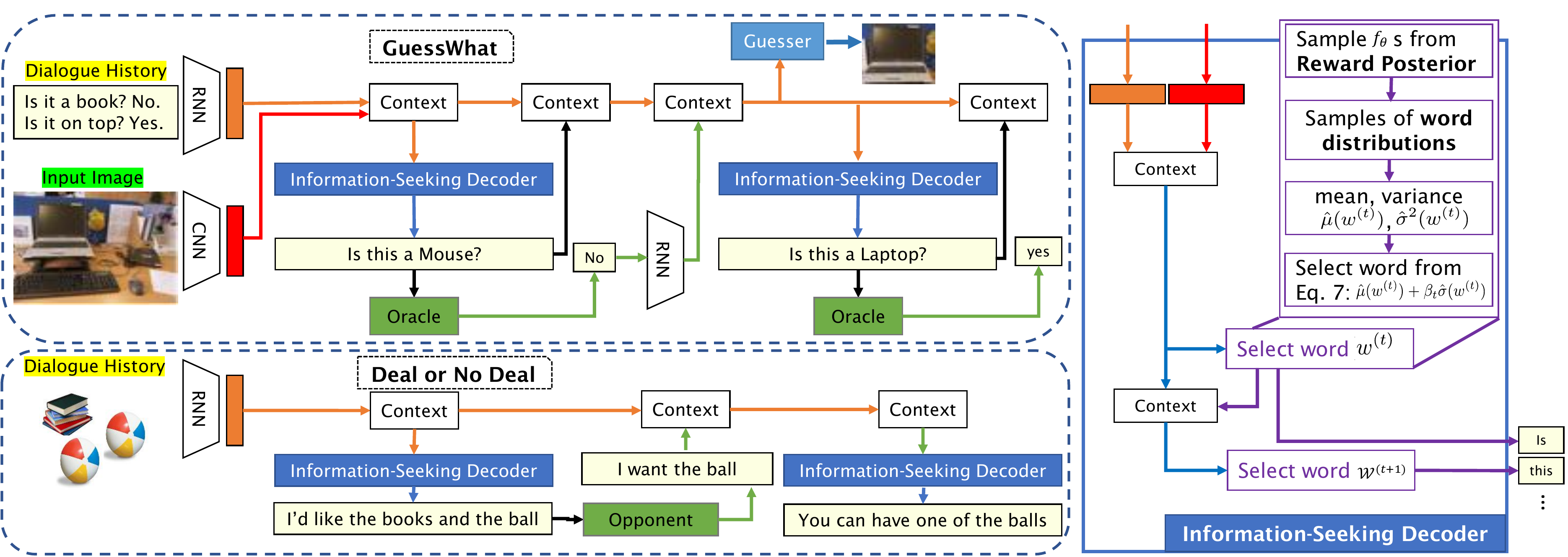}
\caption{The framework in two applications in this paper: we develop a generic information-seeking decoder for dialogue systems.
	Our decoder selects each word optimistically with an upper bound on the reward to maximise the information obtained (details in Sec.~\ref{section:method}). Samples from the reward posterior is taken by applying dropout to the context variable of the RNN
	which provably performs variational inference (see Eq. \ref{eq:posterior} and the Supplements for details).\vspace{-8pt}
}\label{fig:framework}
\end{figure*}

We thus propose an \emph{information-seeking decoder} (see Fig. \ref{fig:framework})
that chooses each word in a question based on its uncertainty about the environment and conditioned
on the history of the conversations. 
We prove this leads to the minimum \emph{expected regret}.
%
An additional benefit of having an accessible estimate of a model's uncertainty is that it allows a more systematic identification of the point at which enough information has been gathered to make the required decision.


We evaluate our model primarily on the well-known collaborative goal oriented visual dialogue problem GuessWhat~\cite{guesswhat_game}.  To demonstrate that it is equally applicable to (non-collaborative) negotiation tasks we also relate its performance on Deal or No Deal~\cite{dealornodeal}.
GuessWhat is a visual dialogue game between two agents in which they cooperate
to identify one of many objects in an image. Deal
or No Deal challenges two players to partition a collection of items
such that each is assigned to one only player. In contrast to
GuessWhat, this game is semi-cooperative in that one player can win
more than their counterpart. Our approach significantly outperforms
the baseline on both tasks.
Our framework is summarised in Fig. \ref{fig:framework}. 


Overall, our contributions are fourfold: 
\begin{itemize}
\vspace{-5pt}
\item We propose a Bayesian Deep Learning method for quantifying the uncertainty in the internal representation of a Reinforcement Learning model. 
This is significant as it provides a theoretically sound method for propagating uncertainty to the output space of the model.
\vspace{-5pt}
\item We describe an {uncertainty-aware information-seeking decoder} for goal-oriented conversation that actively formulates questions that will provide the information the agent needs to achieve its objective.
\vspace{-5pt}
\item We devise a method that exploits the confidence of the
predictor as a measure to indicate if the model has enough information
to produce an accurate output. We show this approach is effective
and leads to fewer rounds of conversation for a goal to be achieved.
\vspace{-5pt}
\item We show that in both visual and textual  dialogue challenges, whether cooperation or adversarial behaviour is desired, our approach outperforms the baselines. To the best of our knowledge, this is the first approach that works well across domains and tasks.
\end{itemize}

\section{Related Work}

\vspace{-3pt}
\paragraph{Goal-oriented dialogue}

Dialogue generation \cite{li2016deep,li2017adversarial,sordoni2015neural}
has been studied for many years in the NLP literature, and has many applications. Dialogue
generation is typically viewed as a Seq2Seq
problem, or formulated as a statistical machine translation problem
\cite{ritter2011data,sordoni2015neural}. Recently, dialogue systems
have been extended to the visual domain. For example, Das \etal \cite{das2016visual}
proposed a visual dialogue task that allows a machine to chat with a
human about the content of a given image.
Goal-oriented dialogue requires the agent understand a user request and complete
a related task with a clear goal within a limited number of 
turns. Early goal-oriented dialogue systems \cite{wang2013simple,young2013pomdp}
model conversation as partially observable Markov Decision Processes (MDP)
with many hand-crafted features for the state and action space representations,
which restrict their usage to narrow domains. Bordes \etal \cite{bordes2016learning}
propose a goal-oriented dialogue test-bed that requires a user chat
with a bot to book a table at a restaurant. 
In visual goal-oriented dialogue De Vries \etal \cite{guesswhat_game} propose a
guess-what game style dataset, where one person asks questions about
an image to guess which object has been selected, and the second person
answers questions as yes/no/NA.

\vspace{-10pt}
\paragraph{RL in dialogue generation}
Reinforcement learning (RL) has been applied in many dialogue settings.
Li \etal \cite{li2016deep} simulate two virtual agents and hand-craft
three rewards to train the response generation model. Recently, some works \cite{asghar2016online,su2016continuously}
make an effort to integrate the Seq2Seq model and RL. RL has also
been widely used to improve dialogue managers, which manage transitions
between dialogue states \cite{pietquin2011sample,singh2002optimizing}.
In visual dialogue, Das \etal \cite{das2016visual} use reinforcement
learning to improve cooperative bot-bot dialogues, and Wu \etal \cite{wu2017you}
combine reinforcement learning and generative adversarial networks
(GANs) to generate more human-like visual dialogues. In \cite{das2017learning}, Das \etal introduce a reinforcement learning mechanism for visual dialogue generation. They establish two RL agents corresponding to question and answer generation respectively, to finally locate an unseen image from a set of images. The question agent predicts the feature representation of the image and the reward function is given by measuring how close the representation is compared to the true feature.


\vspace{-12pt}
\paragraph{Uncertainty}

There are typically two sources of uncertainty to be considered: \emph{Aleatoric}
and \emph{Epistemic} \cite{what_uncertainty}.
The former addresses the noise inherent in the observation while the later
captures our ignorance about which model generated our data. Both
sources can be captured with \emph{Bayesian deep learning}
approaches, where a prior distribution over the model weights is considered.
However, performing Bayesian inference on a deep
neural network is challenging and computationally expensive. One simple
technique that has recently gained attention is to use Monte Carlo
dropout sampling which places a Bernoulli distribution over network
weights~\cite{bayes_dropout,rnn_dropout}. 

Most recently, Lipton \etal \cite{lipton2017bbq} proposed a Bayes-by-Backprop Q-network (BBQ-network) to approximate the Q-function and the uncertainty in its approximation. It encourages a dialogue agent to explore state-action regions in which the agent is relatively uncertain in its action selection. However, the BBQ-network only uses Thompson sampling to model the distribution of rewards for the words in a Bayesian manner and ignores the uncertainty in the estimators. This can lead to very uncertain decisions about confident actions or vice versa. Our method, on the other hand, models both the uncertainty in the actions (i.e. word choices), and the estimators, by directly incorporating the variance in the sampling procedure. We also provide a theoretical justification for the selection which is guaranteed to minimise regret.

\section{Goal-oriented Dialogue Systems}

We ground our goal-oriented dialogue problem as an interactive game between two agents for a collection of items.
The items are either 1) multiple objects in an image for one agent to identify by asking the other questions, or 2) objects for the agents to split by negotiation. 
Conditioned on this game, once enough information
is gathered, a \emph{Guesser} takes the dialogue history and predicts
the goal. The game is a success when the goal is achieved. The game
between these two agents effectively simulates real natural language based conversation
to achieve a particular goal, e.g. uncovering an unknown object, or an agreed split.

Each game is defined as a tuple $(I,\hist,\Y,\ystar)$, where $I$ is the
observed collection, $\hist$ is the dialogue with $T_{\text{dialogue}}$ rounds of
conversation pairs $(\bactions_j,\bactions'_j)_{j=1}^{\Tdialogue}$ and 
$\bactions_j=(\actiont)_{t=1}^{M_{j}}$ is a sequence
of ${M_{j}}$ tokens $\actiont$ with a 
a predefined vocabulary $V$, and $\bactions'_j$ is the response.  
$\Y=(\y_{n})_{n=1}^{N_\y}$
is the list of objects, where $N_\y$ is the number of candidate objects in the collection.
$\ystar$ is the target or a list of targets. In the GuessWhat game \cite{guesswhat_game}, $\ystar$ is a target object that the dialogue refers to. In the Deal or No Deal~\cite{dealornodeal}, it is a list of target objects that the negotiator agent is interested in.


Given an input collection $I$, an initial statement $\bactions_{1}$ is generated
by sampling from the model until the stop token is encountered. Then
the counterpart agent receives the statement $\bactions_{1}$, and generates
the answer $\bactions_{1}'$, the pair $(\bactions_{1},\bactions_{1}')$ is appended to the
dialogue history. We repeat this loop until the end of dialogue token
is sampled, or the number of questions reaches the maximum. Finally,
the \textit{Guesser} takes the whole dialogue $D$ and the object list $\Y$
as inputs to predict the goal. We consider the goal reached if $\ystar$
is selected.

\subsection{RL for Dialogue Generation}

\label{section:reinforcement_learning}


We model dialogue generation as a Markov Decision Process (MDP) to be solved by using
a reinforcement learning (RL) agent~\cite{rl_intro}. The agent interacts
with the environment over a sequence of discrete steps in which we
have the dialogue generated based on the collection $I$ at time step
$t$ in round $T$, the state of agent with the history of conversation pairs and
the tokens of current question generated so far: 
$S_{t}=\big(I,(\bactions_j,\bactions_j')_{j=1}^{T-1},(\actiont_T)_{t=1}^{m}\big)$,
where $t=\sum_{k=1}^{T-1}M_{k}+m$. The action of
agent is to choose the subsequent token $\action_T^{(t+1)}$ from
the vocabulary $V$ (we drop $T$ for brevity). Depending on the 
action the agent takes, the
transition between two states falls into one of the following:

1) $\action^{t+1}=$ end of statement: The current statement is
finished, it is the other agent's turn.

2) $\action^{t+1}=$ end of dialogue: The dialogue is finished, the
\textit{Guesser} selects the output from list $O$.

3) Otherwise, the newly generated token $\action^{t+1}$ is appended 
to the current statement, the next state 
$S_{t+1}=\big(I,(\bactions_j,\bactions_j')_{j=1}^{T-1},(\actiont_T)_{T=1}^{m+1}\big)$.

The maximum length of a statement $\bactions_{j}$ is $M_{max}$, and the maximum number of
rounds in a dialogue is $T_{\text{dialogue}}$. Therefore, the number
of time steps $t$ of any dialogue are $t\leq M_{max}*T_{\text{dialogue}}$.
We use the stochastic policy $\pi_{\btheta}(\action|S)$,
where $\btheta$ represents the parameters of the deep neural network that produces the probability distributions for each state.
The goal of the policy learning is to estimate the parameter $\btheta$.
At the end of the dialogue, a decision about the unknown goal is made
for which a reward is given by the environment. RL seeks to maximise
the expected reward.

After a complete dialogue is generated, we update the RL agent's parameters
based on the outcome of the dialogue. Let $\rw$ be the reward for achieving the goal after completing the dialogue, $\gamma$ be a discount factor,
and $b$ be a bias function estimating the running average of the
completed dialogue rewards so far\footnote{This bias function reduces the variance of the estimator.}.
Let future reward $R$ for an action $\actiont$ be
$R(\actiont)=\mathbb{E}\big[\sum_{i=0}^{\infty}\gamma^{i}(r_{\action^{t+i}}-b(\action^{t+i}))\big]$
where expectation is with respect to the policy $\pi$. The parameters
of this model comprising of the policy and the bias function are then
optimised using gradient policy theorem~\cite{policy_grad} and REINFORCE
\cite{reinforce}. The policy determines how a statement is made in a dialogue system.
Note that at each step there is an estimation of
the reward (which is never directly observed) for each word in the
RL and the observable reward is only given to the complete dialogue.
Upon receiving the reward for the complete dialogue the parameters are
accordingly updated. Utilising this estimation of the reward at each
stage and a particular choice of the word strategy, a sequence of
words is generated.

In the subsequent section we discuss a particular
strategy that utilises the {uncertainty} in the policy (model) and seeks
to provide a better approach for exploration of the space of possible
dialogues. Moreover,  since REINFORCE is a Monte Carlo estimate that is known to have a high variance, there is an additional source of uncertainty in evaluation of the expected rewards. As such, it is essential to consider uncertainty in policies manifesting in word choices.

\subsection{Information-seeking Decoder}
\label{section:method}
The decoder's objective is, given the dialogue thus far, to choose the subsequent
word such that the resulting 
response
is most ``informative''. To
that end, we assume there is an underlying reward for each word $\rw$
at step $t$ that we seek to uncover by exploring the space
of actions (tokens in the vocabulary). A common practice is to model this value
as the output of a \emph{deterministic} function $\func:V\to\mathbb{R}$
parameterised by $\boldsymbol{\theta}$ such as a neural network for
sequential problems (e.g. LSTMs \cite{lstms} or GRUs \cite{grus}). To select the subsequent action
using this function one can greedily select the action with highest
value or sample from a softmax (categorical distribution) built from
this function.%

However, this approach does not account for the uncertainty in the
prediction of the reward $\rw$. This uncertainty has two main sources,
(1) model uncertainty which is due to the imperfections in the parameters
and (2) prediction uncertainty which is due to the lack of information
about each action and its consequence. We choose a prior for the parameters
and update them with the likelihood of the dialogue observations to obtain the posterior 
distribution in a Bayesian manner. The posterior at round $T$ is:
{\small\begin{equation}
p(\btheta|\brewards_{t},\histt, I) = 
\frac{1}{Z} \prod_{t}p(\action^{(t)}|(\bactions_j,\bactions_j')_{j=1}^{T-1},\brewards_{t},I,f_{\btheta})p(\btheta)
\label{eq:posterior}
\end{equation}} 
\noindent where $Z$ is the normaliser and $p(\btheta)$ is a 
prior for the parameters. Here, 
 $\brewards_{t}={r_{\action^{(1)}},\ldots,r_{\action^{(t)}}}$ is the 
 set of rewards collected up to step $t$ in the dialogue.
This formulation has a self-regularising behaviour that, unlike likelihood maximisation,
is less susceptible to a local optima and performs better in practice. The predictive distribution 
of rewards from which each word is chosen becomes\footnote{An alternative view is that we model $f_{\btheta}$ as a stochastic function and choose words accounting for their uncertainty. $f_{\btheta}$ is fully realised by its parameters $\btheta$, hence we use the uncertainty in the functional and the parameters interchangeably.}:
{\begin{align}
& p(r_{\action^{(t+1)}}|\action^{(t+1)},\brewards_{t},\histt, I)\nonumber \\
& \quad=\int p(r_{\action^{(t+1)}}|\action^{(t+1)},f_{\btheta})p({\btheta}|\brewards_{t},\histt, I)d{\btheta}\label{eq:rewards_sample} \\
&\quad\approx\frac{1}{N}\sum_{i=1}^{N}p(r_{\action^{(t+1)}}|\action^{(t+1)},f_{\btheta}^{(i)}),\,\,f_{\btheta}^{(i)}\sim p({\btheta}|\brewards_{t},\histt,I)\nonumber
\end{align}}
\noindent where $N$ is the number of samples for the Monte Carlo estimation
of the integral and
{\begin{equation}
\begin{aligned} & p(r_{\action^{(t+1)}}|\action^{(t+1)},f_{\btheta})=\text{softmax}(f_{\btheta}(\action_1^{(t+1)}),\ldots,\action_{|V|}^{(t+1)})
\end{aligned}
\end{equation}} where $|V|$ is the size of the dictionary.
However, the posterior $p(\btheta|\brewards_{t},\histt, I)$ in Eq.\ref{eq:posterior} does not have a closed-form solution. Thus, we resort to variational inference~\cite{variational_infer}, the details of which is provided in the Supplements. In a nutshell, inspired by \cite{bayes_dropout,rnn_dropout} we show 
that the posterior is approximated by a particular mixture model which is equivalent to performing typical MAP with dropout regularisation for dialogue generation. 
Further, the Monte Carlo estimate in 
Eq.\ref{eq:rewards_sample} is efficiently computed by applying dropout $N$ times in the RNN network (note we take $N$ context variables in Fig. \ref{fig:framework}).
%
Hence, the mean and variance of the rewards computed from the posterior are: 
\begin{align}
 & \mean=\frac{1}{N}\sum_{i=1}^{N}f_{\btheta}^{(i)}(\actiont)\label{eq:mean_std}\\
 & \var=\frac{1}{N}\sum_{i=1}^{N}\left(f_{\btheta}^{(i)}(\actiont)-\mu(\actiont)\right)^{2}+\tau^{-1}
\end{align}
where $\tau$ is the precision parameter. Using Chebyshev's inequality, we have:
\begin{align}
& p\left(\big|\func-\mean\big|<\beta_{t}\std\right) \!& \geq & \,\,\,1-\frac{1}{\beta_{t}^{2}}
\vspace{-3pt}
\end{align}
which means for $\beta_{t}>0$, with high probability we have $\big|\func-\mean\big|<\beta_{t}\std$
for a random function $\func$. Hence we have an \emph{upper bound}
on the random function $f_{\btheta}$ with high probability:
\begin{eqnarray}
\vspace{-3pt}
\func & < & \mean+\beta_{t}\std\label{eq:upper_bound}
\vspace{-3pt}
\end{eqnarray}
Selecting an action (word) with this upper bound both accounts for
the estimation of the high--reward values by $\mean$ and the
uncertainty in this estimation for the given word $\std$. In this
bound, $\beta_{t}$ controls how much the uncertainty is taken into
account for selecting a word. Furthermore, it is clear that with $\beta_{t} \to 0$
this upper bound approaches greedy selection. In the reinforcement
learning context, this approach 
mediates
the exploration-exploitation
dilemma by changing $\beta_{t}$.

This upper-bound is inspired by the Upper Confidence Bound (UCB) which
is popular in multi-armed bandit problems \cite{ucb}. A similar
upper bound for Gaussian processes was proposed in \cite{gp_ucb}.
However, this bound for neural networks, in particular for dialogues
systems, is novel.

\vspace{-8pt}
\paragraph{Expected Regret and Information}

For a dialogue agent, a metric for evaluating performance is cumulative
regret, that is the loss due to not knowing the best word to choose
at a given time. Suppose the best action at round $t$ is $\actiont_{*}$
for our choice $\actiont$ , we incur instantaneous expected regret,
$\regrett=\mathbb{E}_{f_{\btheta}}\big[f_{\btheta}(\actiont_{*})-f_{\btheta}(\actiont)\big]$.
The cumulative regret $\Regrett$ after $T$ rounds is the sum of
instantaneous expected regrets: $\Regrett=\sum_{t=1}^{T}\regrett$.
Note that neither $\regrett$ nor $\Regrett$ are ever revealed during
dialogues generation. Our expected regret at each round is bounded as
{\small\begin{align}
\vspace{-5pt}
& \regrett < \mathbb{E}_{f_{\btheta}}\big[\mean+\beta_{t}\std-f_{\btheta}(\actiont)\big] < 2\beta_{t}\std
\vspace{-5pt}
\end{align}}
where the first inequality is due to $f_{\btheta}(\actiont_{*})<\mean+\beta_{t}\std$
and the second one is because $\big|\func-\mean\big|<\beta_{t}\std$,
then $-\func<-\mean+\beta_{t}\std$. Therefore, we have
{\small
\begin{eqnarray}
\vspace{-5pt}
\Regrett & < & 2\sum_{t}\beta_{t}\std
\end{eqnarray}}
As such, the expected regret for each word selected is bounded by
the standard deviation of the predicted reward. When we choose words
with high standard deviation, we actively seek to gain more information
about the uncertain words to effectively reduce our expected regret.

Further, let's assume the predictive distribution is near Gaussian
with mean and variance $\mean$, $\var$ (which considering the central
limit theorem is natural).  The entropy is then $\frac{1}{2}\log(2\pi e\var)$.
Hence, selecting actions with higher uncertainty is also justified
from an information-theoretic perspective as means of selecting informative
words. In a dialogue system, when uttering a sentence with length $T$
the information we can obtain is at most (using the union bound) $\frac{1}{2}\sum_{t}^{T}\log(2\pi e\var)$.
An alternative to using the approach in Eq. \ref{eq:upper_bound}
is to choose the words with highest entropy (the most
informative words). However, that is an extreme case that will lead
the RL algorithm to continuously explore the dialogue space.

\begin{algorithm}[t]
{\footnotesize
\begin{algorithmic}[1]
\For{Each update}
\State \# Generate trajectories
\For{$k = 1 \text{ to } K$} \Comment{Select $K$ of the objects/items}
\State Pick target objects $o_k^* \in O_k$
\State Set $\histt$ to initial input collection
\For{$j = 1 \text{ to } \Tdialogue$} \Comment{Generate $(\bactions_j, \bactions_j')$ pairs}
\While{$\action^{(t+1)}$ not <stop>}
\State Sample $f^{(n)}_{\btheta}\sim p(f_{\btheta}|\bactions_{j},\brewards_{t})$, $n=1,\ldots,N$
\State Set $\mean, \std$ from $f^{(n)}_{\btheta}$
\State $\action^{(t+1)}=\arg\max_\action\,\, \hat{\mu}(\action)+\beta_t\hat{\sigma}^2(\action)$
\EndWhile
\State $\bactions_{j}' = \text{SuperviseAgent}(\bactions_{j}, \histt)$
\If{$\stopdialoguetoken \in \bactions_{j}$ or $H(o_{t+1}|\histt)\leq\eta$ } \Comment{Sec.~\ref{section:stop}}
\State delete $(\bactions_{j},\bactions_{j}')$ and break;
\Else
\State append $\action^{(t+1)}$ to $\bactions_{j}$
\EndIf
\State append $(\bactions_{j},\bactions_{j}')$ to $\histt$
\EndFor
\State $o_k = \argmax_o\, p(o|\histt)$ \Comment{Predict the goal}
\State $\text{reward}=\begin{cases} 1 & \text{If } o_k=o_k^*\\     0 & \text{Otherwise}\\
\end{cases}$
\EndFor
\State Evaluate policy and update parameters $\btheta$
\EndFor
\end{algorithmic} }
\caption{Training information-seeking dialogues}\label{reinforce}
\label{alg}
\end{algorithm}

\subsection{\vspace{-1mm}Stopping Dialogue}
\label{section:stop}
\vspace{-1mm}

One of the key challenges in a goal-oriented dialogue system is to identify the point at which the agent has sufficient information to make the required decision.  
We specified above that the probability 
of the unknown goal given the dialogue thus far is $p(o_{t+1}|\histt)$.
The uncertainty in this measure reflects the agent's confidence in its prediction, and thus provides a natural
measure for the stopping criteria of the conversation. Intuitively,
the agent stops when it feels confident in its prediction of the goal.
Hence, we have $H(o_{t+1}|\histt)\leq\eta$
where $H$ is the entropy and $\eta$ is an appropriately chosen hyper-parameter
for the confidence. When $\eta$ is larger, we allow for less confident
predictions leading to shorter dialogues. See Alg. \ref{alg} for the full algorithm.

\section{\vspace{-2mm}Experiments}

To evaluate the performance of the proposed approach we conducted experiments
on two different goal-oriented dialogue tasks: GuessWhat~\cite{guesswhat_game} and Deal or No Deal~\cite{dealornodeal}. 
Our approach outperforms the baseline in both cases. In
both experiments we pre-train the networks using the supervised
model 
and refine using reinforcement learning.
To that end, we employ a two stage algorithm in which
we learn to imitate the human dialogue behaviour in a supervised learning
task and subsequently fine-tune for better generalisation and goal
discovery using reinforcement learning. 
 In both experiments, our decoder takes 
the history of the dialogue in addition to input collection (e.g. an image)
and, guided by the uncertainty of each word,
produces a question.
Similar two-stage approaches
are taken in \cite{visdial_rl,guesswhat_game,dealornodeal}. Without
using supervised learning first, the dialogue model may diverge from
human language.

\subsection{\vspace{-2mm}GuessWhat}

In GuessWhat \cite{guesswhat_game} a visually rich image with several objects is shown to two players.
One player selects an object from the image. The task of the other player, the questioner,   is
to locate the unknown object by asking a series of yes/no questions. After enough information is gathered by the
questioner, it then guesses what the selected object was. If the questioner
guesses the correct object the game is successfully concluded. It is
desirable for the questioner to guess the correct answer in as few
rounds of questioning as possible. The dataset includes $155,281$ dialogues of
$821,955$ pairs of question/answers with vocabulary size $11,465$ on
$66,537$ unique images and $134,074$ objects. 

\vspace{-13pt}
\paragraph{Implementation Details}
We follow the same experimental setup as \cite{guesswhat_game} in
which three main components are built: a yes/no answering agent, a guesser
and a questioner. The questioner is a recurrent neural
network (RNN) that produces a sequence of state vectors for a given
input sequence by applying long-short term memory (LSTM)
as a transition function. The output of this LSTM network is the internal
estimate of the reward with size $1024$. To obtain a distribution over tokens, a softmax
is applied to this output.

\begin{table}
\centering%
\resizebox{1\linewidth}{!}{
\begin{tabular}{c|c|c|c|c}
\hline
\multicolumn{5}{c}{{{}New Object}}\tabularnewline
\hline
 & {Sampling} & {{}Greedy} & {{}Beam Search} & {{}Avg. Ques}\tabularnewline
\hline
{{}Supervised}~\cite{guesswhat_game}
 & {{}$41.6$} & {{}$43.5$} & {{}$47.1$} & {{}5}\tabularnewline
{{}RL}~\cite{strub2017end} & {{}$58.5$} & {{}$60.3$} & {{}$60.2$} & {{}5}\tabularnewline
{{}TPG}~\cite{tempered} & {{}$62.6$} & -&- &{{}5} \tabularnewline
\hline
{{}Ours} & {{}$\boldsymbol{61.4}$} & {{}$\boldsymbol{62.1}$} & {{}$\boldsymbol{63.6}$} & {{}5}\tabularnewline

{{}Ours ($\eta=0.05$)} & {{}$58.5$} & {{}$59.5$} & {{}$59.6$} & {{}\textbf{4.2}}\tabularnewline

{{}Ours ($\eta=0.01$)} & {{}$59.8$} & {{}$59.3$} & {{}$60.4$} & {{}4.5}\tabularnewline
\hline
{{}Ours+MN} & {\boldsymbol{$68.3$}} & {\boldsymbol{$69.2$}} & -& {{}5}\tabularnewline
\hline
\hline
\multicolumn{5}{c}{{{}New Image}}\tabularnewline
\hline
{{}Supervised}~\cite{guesswhat_game}
 & {{}$39.2$} & {{}$40.8$} & {{}$44.6$} & {{}5}\tabularnewline
{{}RL}~\cite{strub2017end} & {{}$56.5$} & {{}$58.4$} & {{}$58.4$} & {{}5}\tabularnewline
\hline
{{}Ours } & {{}$\boldsymbol{59.0}$} & {{}$\boldsymbol{59.82}$} & {{}$\boldsymbol{60.6}$} & {{}5}\tabularnewline

{{}Ours ($\eta=0.05$)} & {{}$56.7$} & {{}$56.5$} & {{}$57.3$} & {{}\textbf{4.3}}\tabularnewline

{{}Ours ($\eta=0.01$)} & {{}$58.0$} & {{}$57.5$} & {{}$58.5$} & {{}4.5}\tabularnewline
\hline
{{}Ours+MN} & {\boldsymbol{$66.3$}} & {\boldsymbol{$67.1$}} & - & {{}5}\tabularnewline
\hline \hline
\end{tabular}}
\vspace{2pt}
\caption{Accuracy in identifying the goal object in the GuessWhat dataset (higher is better). The numbers in parentheses is the threshold used in questions for guessing the object in the image. Average number of questions is shown at the last column (lower is better).}
\label{tbl:guesswhat1}
\vspace{-7pt}
\end{table}

The samples of the reward estimate in the questioner are taken utilising
dropout with parameter $0.5$. Subsequently, the upper bound in Eq.~\ref{eq:upper_bound} is calculated to choose words.
For this experiment we set $\beta_t=1$\footnote{We observed marginal performance improvement by using a larger $\beta$ on Guesswhat, despite the additional training overhead.}.

Once the questioner is trained using our information seeking decoder in RL, we take three approaches
to evaluating the performance of the questioner: (1) \textit{sampling} where
the subsequent word is sampled from the multinomial distribution in
the vocabulary, (2) \textit{greedy} where the word with maximum probability
is selected and (3) \textit{beam search} keeping the K-most promising
candidate sequences at each time step (we choose $K=20$ in all
experiments). During training the baseline uses the greedy approach
to select the sequence of words as in \cite{guesswhat_game}.

\vspace{-14pt}
\paragraph{Overall Results}
We compare two cases, labelled \textit{New Object} and \textit{New Image}.  In the former the object sought is new, but the image has been seen previously.  In the latter the image is also previously unseen.
We report the prediction
accuracy for the guessed objects. It is clear that the accuracies
are generally higher for the new objects as they are obtained from
the already seen images.

\begin{figure}[t]
\resizebox{1\linewidth}{!}{
\begin{tabular}[b]{c}
\hline
{\footnotesize{}\hspace{-3mm}}%
\begin{tabular}{>{\raggedright}p{3.5cm}ccc}
{\footnotesize{}is it a person?} & {\footnotesize{}Yes} & \multirow{7}{*}{{\footnotesize{}\hspace{-3mm}\includegraphics[scale=0.15]{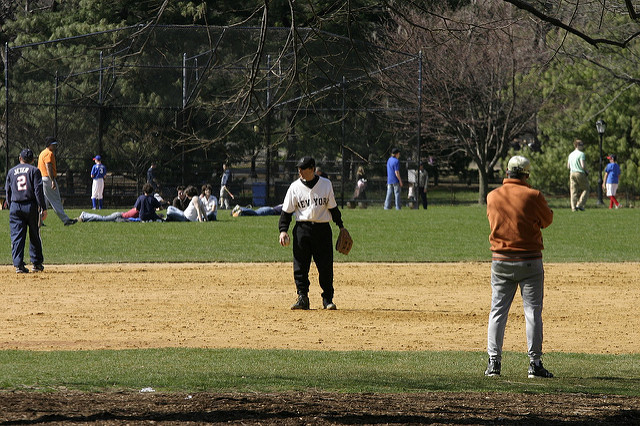}\hspace{-3mm}}} & \multirow{7}{*}{\begin{turn}{90}
{\footnotesize{}Man in Orange}
\end{turn}}\tabularnewline
{\footnotesize{}is he wearing a brown coat?} & {\footnotesize{}No} &  & \tabularnewline
{\footnotesize{}is he wearing a white shirt?} & {\footnotesize{}No} &  & \tabularnewline
{\footnotesize{}is he wearing a blue shirt?} & {\footnotesize{}No} &  & \tabularnewline
{\footnotesize{}are they sitting down?} & {\footnotesize{}No} &  & \tabularnewline
{\footnotesize{}is it the guy in the orange shirt to the left?} & {\footnotesize{}Yes} &  & \tabularnewline
\end{tabular}
\tabularnewline
\hline
\end{tabular}}
\caption{{Sample dialogue from the GuessWhat dataset.
	The agent asks about a brown coat and then
	changes it to orange in anticipation of wrong identification.}
	}
\label{exm_figure}
\vspace{-12pt}
\end{figure}

The results are summarised in Tab. \ref{tbl:guesswhat1}. As shown,
simply applying REINFORCE improves the output of the system significantly,
in particular in the new image case where the generalisation is tested.
This improvement is because the question generator has the chance
to better explore possible questions. Additionally,
the greedy approach outperforms others in the RL baseline in \cite{strub2017end}.
This illustrates that the distribution of the words obtained from the
softmax in the question generator is not very peaked and the difference
between the best and second best word is often small. This indicates that the prediction at test time is very
uncertain and supports our approach.

Since our approach seeks uncertain words, those words are exploited
at training time, which leads to lower variance (a more peaked distribution)
and better performance of the greedy selection. Beam search 
significantly increases performance when we carry out $5$ rounds (as in \cite{guesswhat_game, strub2017end}) of question-answering. This is because the most informative words are selected by our approach
which, combined with the beam-search's mechanism for forward exploration,
leads to better performance.

Note that our approach is generic enough that can be used in combination with other architectures 
(e.g. \cite{answer_questioner_mind, zhang2018goal}). For instance, in Tab.~\ref{tbl:guesswhat1} ``{Ours+MN}'' uses the Memory Network \cite{memorynet} and Attention mechanism \cite{attention} in the Guesser (similar to that of \cite{tempered}) 
which leads to better question generation. 
Fig. \ref{exm_figure} shows one example produced by our dialogue generator. More examples can be found in the supplements.

\begin{figure}
	\vspace{-10pt}
	\centering\includegraphics[width=0.68\columnwidth]{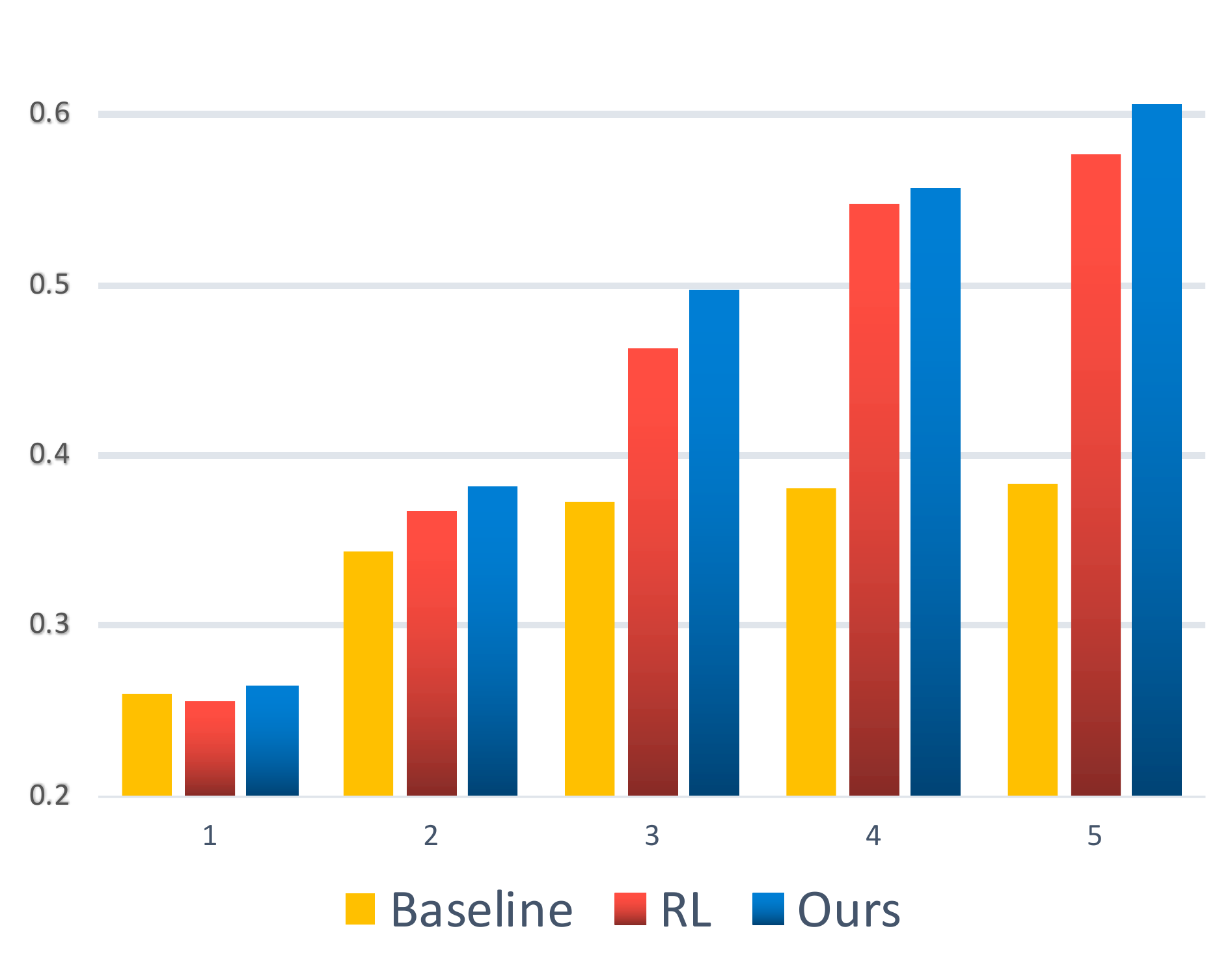}
	\caption{\small{The proportion of dialogues successful in identifying 
			the goal object at each round in GuessWhat.}
	\vspace{-12pt}}
	\label{proportion}
\end{figure}

\vspace{-13pt}
\paragraph{Ablation Study on Early Stopping}
In goal-oriented dialogue systems, it is desirable to make a decision as
soon as possible. 
In this experiment, we control the dialogue length by changing the 
threshold $\eta$ (see Sec. \ref{section:stop} for more details). 
When $\eta$ is larger, we accept less confident predictions leading 
to shorter dialogues. As shown in the Tab. \ref{tbl:guesswhat1}, our 
models achieve a comparable performance to the baseline even using 
shorter rounds of question answering. The Fig. \ref {proportion} 
shows the proportion of dialogues successful in identifying the goal 
object at each round. Our model achieves higher accuracy even in the 
earlier rounds, e.g. at the round three.

\vspace{-13pt}
\paragraph{Human Study}
To evaluate how well humans can
guess the target object based on the questions generated by
our models, we conduct a human study. Following \cite{zhang2018goal}, we show human subjects 50 images with
generated question-answer pairs from our model, and let them guess the objects. We ask three
human subjects to play on the same split and the game is
recognised as successful if at least two of them give the
right answer. In our experiment, the average performance of humans was 79\% compared to 52\% and 70\% for the supervised \cite{guesswhat_game} and RL~\cite{strub2017end} models. We are even better than a model proposed in \cite{zhang2018goal} (76\%), which has three complex hand-crafted rewards. These results indicate that our agent can provide more useful information that can benefit a human in achieving the final goal.


\subsection{\vspace{-2mm}Deal or No Deal}
\label{sec:DealOrNoDeal}

\begin{table}[b]
\vspace{-11pt}
\resizebox{1\linewidth}{!}{
\hspace{-7mm}\begin{tabular}{lr|c|c|c}
\cline{2-5}
 &  & {\small{}Score} & {\small{}\% Agreed} & {\small{}\% Selection}\tabularnewline
\cline{2-5}
& {{\begin{turn}{90}
\noindent\hspace{-11mm}{
\,\,\,{\footnotesize{Baseline}}
}\end{turn}}}\hspace{5mm}
{\footnotesize{}Supervised \cite{dealornodeal}} & {\footnotesize{}$5.4$ vs. $5.5$} & {\footnotesize{}$87.9$} & {\footnotesize{}$50.78$ vs $49.23$} \tabularnewline

 & {\footnotesize{}RL \cite{dealornodeal}} & {\footnotesize{}$7.1$ vs. $4.2$} & {\footnotesize{}$89.9$} & {\footnotesize{}$55.81$ vs $44.19$}\tabularnewline

 & {\footnotesize{}RL+Rollouts \cite{dealornodeal}} & {\footnotesize{}$8.3$ vs. $4.2$} & {\footnotesize{}$94.4$} & {\footnotesize{}$60.02$ vs $39.98$}\tabularnewline
\cline{2-5}
 & {\footnotesize{}$\beta_{t}=1$} & {\footnotesize{}$8.09$ vs $4.08$} & {\footnotesize{}$92.02$} & {\footnotesize{}$77.13$ vs $22.87$}\tabularnewline

 & {{\begin{turn}{90}
 \noindent\hspace{-12mm}{
 \,\,\,\,\,\,\,\,\,{Ours}}
 \end{turn}}}\hspace{14mm}{\footnotesize{}$\beta_{t}=10$} & {\footnotesize{}$8.27$ vs $4.23$} & {\footnotesize{}$94.79$} & {\footnotesize{}$88.56$ vs $11.44$ }\tabularnewline

 & {\footnotesize{}$\beta_{t}=1000$} & {\footnotesize{}$8.21$ vs $4.33$} & {\footnotesize{}$94.65$} & {\footnotesize{}$87.05$ vs $12.95$}\tabularnewline

 &
 {\footnotesize{}$\beta_{t}=10$+Rollouts} & {\footnotesize{}$\boldsymbol{8.58}$ vs $4.13$} & {\footnotesize{}$\boldsymbol{95.75}$} & {\footnotesize{}$\boldsymbol{93.62}$ vs $6.38$}\tabularnewline
\cline{2-5}
\end{tabular}}
\vspace{2pt}
\caption{\small Prioritising words with greater uncertainty leads to better performance in negotiations.`$\%$ Selection' represents the percentage of trials in which the final decision is made by each agent.\label{tbl:results_deal_nodeal}\vspace{-5pt}}
\end{table}

Here two agents receive
a collection of items, and are instructed to divide them so that each
item is assigned to one agent. This problem is, unlike the GuessWhat
game, semi-cooperative game in that the goals are adversarial.
Each agent's goal is to maximise its own rewards which may be in direct
contradiction with its opponents goals.

Each item has a different random non-negative value for each agent.
These random values are constrained so that: (1)~the sum of values
for all items for each agent is 10; (2)~each item has a non-zero value
for at least one agent; and (3)~there are items with non-zero value
for both agents. These constraints are to ensure both agents cannot
receive a maximum score, and that no item is worthless to both agents.
After 10 turns, agents are
given the option to complete the negotiation with no agreement, which is
worth 0 points to each. There are 3 item types (\emph{books},
\emph{hats}, \emph{balls}) in the dataset and between 5 and 7 total
items in the collection.

\vspace{-15pt}
\paragraph{Implementation Details}
The supervised learning model comprises 4 recurrent neural
networks implemented as GRUs. The agent's input goal
is encoded as the hidden state of a GRU with size $64$. The tokens are generated
by sampling from the distribution of tokens. 
Simple maximum likelihood often leads to accepting
an offer because it is more often 
than proposing a counter offer.
To remedy this problem, similar to the previous GuessWhat experiment, we perform
goal-oriented reinforcement learning to fine-tune the model. In addition,
following \cite{dealornodeal} we experimented with rollouts.  That
is, considering the future expected reward in the subsequent dialogue, which is
similar to the beam search in the previous experiment. 

\vspace{-15pt}
\paragraph{Results \& Ablation Analysis}
Results are shown in Tab.~\ref{tbl:results_deal_nodeal}. We report
the average reward for each agent and the percentage of agreed upon
negotiations. We see that our approach significantly outperforms the
baseline RL. This is due to the information-seeking behaviour of our
approach that leads to the agent learning to perform better negotiations
and achieve agreements when the deals are acceptable.

We also evaluate the influence of $\beta_t$ (in Eq. \ref{eq:upper_bound}), which controls how much the uncertainty is taken into account in selecting a word, in turn controling the extent of exploration in dialogue generation. 
Increasing $\beta_t$ leads to more exploration and more confidence in the actions at the 
expense of later convergence. From Tab.~\ref{tbl:results_deal_nodeal}, 
we can see that a larger $\beta_t$ leads to better performance. We also observed that 
if $\beta_t$ is too high, say 1000, it diminishes performance as the agent continues exploring
(by uttering risky statements that may lead to better understanding of the agent's counterpart at the cost of losing the deal)
rather than exploiting its knowledge about the best word choices at each step of the negotiation.

Tab. \ref{tbl:negotation_example} shows examples of the negotiations
generated using our model. The baseline model sometimes refuses an
option that could lead to a desirable deal. She learns to be forceful
in some cases, and consistent. This is because our model uncovers that by
taking this risk, the counterpart may change his strategy. This is in part due to
the fact that the supervised case the agent is willing to compromise quickly
and our approach exploits that. This is achieved by repeating
the same proposition by our model. Furthermore,
our model learns to give her counterpart an option to give him a false
sense of control over the negotiation, thus deceiving him.
While she seems to have given-up in favour of the other's benefit,
she enforces her choice and is consistent.

\begin{table}[t]
\begin{centering}
\resizebox{\linewidth}{!}{
{\footnotesize{}}
\begin{tabular}{|>{\raggedright}p{3.5cm}||>{\raggedright}p{4.3cm}|}
\hline
\multicolumn{2}{|l|}{{\footnotesize{}Alice : book=(2, 0) hat=(2, 5) ball=(1, 0)}}\tabularnewline
\multicolumn{2}{|l|}{{\footnotesize{}Bob : book=(2, 2) hat=(2, 2) ball=(1, 2)}}\tabularnewline
\hline
{\footnotesize{}Our Approach vs Baseline} & {\footnotesize{}Baseline vs Baseline}\tabularnewline
\hline
\textbf{\footnotesize{}Alice}{\footnotesize{}: i would like the hats
and the books.}{\footnotesize\par}

\textbf{\footnotesize{}Bob}{\footnotesize{}: i need the hats and the
books.}{\footnotesize\par}

\textbf{\footnotesize{}Alice}{\footnotesize{}: you can have the ball
if i can have the rest}{\footnotesize\par}

\textbf{\footnotesize{}Bob}{\footnotesize{}: ok deal} & \textbf{\footnotesize{}Alice}{\footnotesize{}: i'd like the hats and
the ball. }{\footnotesize\par}

\textbf{\footnotesize{}Bob}{\footnotesize{}: you can have the ball
, but i need the hats and the books . }{\footnotesize\par}

\textbf{\footnotesize{}Alice}{\footnotesize{}: i need the hats and
a book .}{\footnotesize\par}

\textbf{\footnotesize{}Bob}{\footnotesize{}: no deal . i can give
you the ball and both books}\tabularnewline
\hline
{\footnotesize{}Alice: }\textbf{\footnotesize{}10}{\footnotesize{}
, Bob: }\textbf{\footnotesize{}2}{\footnotesize{} points} & {\footnotesize{}Alice: }\textbf{\footnotesize{}0}{\footnotesize{}
, Bob: }\textbf{\footnotesize{}4}{\footnotesize{} points}\tabularnewline
\hline
\multicolumn{2}{|l|}{{\footnotesize{}Alice : book=(1, 0) hat=(1, 7) ball=(3, 1) }}\tabularnewline
\multicolumn{2}{|l|}{{\footnotesize{}Bob : book=(1, 9) hat=(1, 1) ball=(3, 0)}}\tabularnewline
\hline
\textbf{\footnotesize{}Bob}{\footnotesize{}: i would like the book
and the hat .}{\footnotesize\par}

\textbf{\footnotesize{}Alice}{\footnotesize{} : you can have the book
if i can have the rest }{\footnotesize\par}

\textbf{\footnotesize{}Bob}{\footnotesize{}: ok , deal} & \textbf{\footnotesize{}Bob}{\footnotesize{}: i want the book and 2
balls}{\footnotesize\par}

\textbf{\footnotesize{}Alice}{\footnotesize{}: i need the hat and
the balls}{\footnotesize\par}

\textbf{\footnotesize{}Bob}{\footnotesize{}: i need the book and one
ball}{\footnotesize\par}

\textbf{\footnotesize{}Alice}{\footnotesize{}: how about i take the
hat and 1 ball?}{\footnotesize\par}

\textbf{\footnotesize{}Bob}{\footnotesize{}: sorry i cant make a deal
without the book }{\footnotesize\par}

\textbf{\footnotesize{}Alice}{\footnotesize{}: then we will need the
hat and the book}\tabularnewline
\hline
{\footnotesize{Alice: }}\textbf{\footnotesize{10}}{\footnotesize{, Bob: }}\textbf{\footnotesize{9}}{\footnotesize{ points}} &
{\footnotesize{Alice: }}\textbf{\footnotesize{0}}{\footnotesize{}
(7)$^*$, Bob: }\textbf{\footnotesize{0}}\footnotesize{ points}
\tabularnewline
\hline
\end{tabular}{\footnotesize\par}}
\par\end{centering}
\vspace{2pt}
\caption{\small Samples from the negotiation experiments: Our approach is Alice and Bob is the baseline. $^*$ is the potential reward.}
\label{tbl:negotation_example}
\vspace{-13pt}
\end{table}

\vspace{-1mm}
\section{Conclusion}

One of the primary limitations of current goal-directed dialogue systems is their limited ability to identify the information required to achieve their goal, and the steps required to obtain it.  This limitation inherent in any reinforcement learning-based system that needs to learn to acquire the information required to achieve a goal.
We have described a simple extension to reinforcement learning that overcomes this limitation, and enables an agent to select the action that is most likely to provide the information required to meet their objective.  The selection process is simple, and controllable, and minimises the expected regret.  It also enables a principled approach to identifying the appropriate point at which to stop seeking more information, and act.

The approach we propose is based on a principled Bayesian formulation of the uncertainty in both the internal state of the model, and the process used to select actions using this state information.
We have demonstrated the performance of the approach when applied to generating goal-oriented dialogue, which is one of the more complex problems in its class due to the generality of the actions involved (natural language), and the need to adapt to the unknown intentions of the other participant.  
The proposed approach none the less outperforms the comparable benchmarks.

{\small
\bibliographystyle{ieee}
\bibliography{lib}
}

\clearpage


\section*{Supplementary material}

# 1. Appendix

**Approximate Variational Inference in Bayesian Neural Networks**

In this section we consider the distribution of weight matrices that parameterises our functions. This is the posterior over the weights given our observables, dialogue history, image and the past rewards (indicated by the existence of a word in a sentence), $D_t, I, \mathcal{R}_t$:

$$p(\boldsymbol{\theta}|\mathcal{R}_t, D_t, I) = \frac{1}{Z} \prod_t p(w^{(t)}|D_t, \mathcal{R}_t, I, f_{\boldsymbol{\theta}}) p(\boldsymbol{\theta})$$

This posterior defines the distribution of the parameters given the current known variables. This posterior is not tractable in general, and we consider using variational inference to approximate it [10]. We utilise an alternative distribution $q(\boldsymbol{\theta})$ that approximates $p(\boldsymbol{\theta}|\mathcal{R}_t, D_t, I)$ by minimise the KL divergence between these two distributions:

$$\text{KL}\big(q(\boldsymbol{\theta})||p(\boldsymbol{\theta}|D_t, I, \mathcal{R}_t)\big) \tag{1}$$

$$\propto - \int q(\boldsymbol{\theta}) \log p(\mathcal{R}_t|D_t, I, f_{\boldsymbol{\theta}}) \mathrm{d}\boldsymbol{\theta} + \text{KL}(q(\boldsymbol{\theta})||p(\boldsymbol{\theta}))$$

$$= -\sum_{j=1}^{T-1} \int q(\boldsymbol{\theta}) \log p(w^{(t)}|(\mathcal{W}_j, \mathcal{W}'_j), \mathcal{R}_t, I, f_{\boldsymbol{\theta}}) \mathrm{d}\boldsymbol{\theta} + \text{KL}(q(\boldsymbol{\theta})||p(\boldsymbol{\theta})). \tag{2}$$

Following [11] we define our approximating distribution to factorise over the weight matrices and their rows in $\boldsymbol{\theta}$. Every element of the parameter matrix of the approximating distribution is:

$$q(\boldsymbol{\theta}_k) = \eta \mathcal{N}(\boldsymbol{\theta}_k; \mathbf{0}, \sigma^2 I) + (1-\eta) \mathcal{N}(\boldsymbol{\theta}_k; \mathbf{m}_k, \sigma^2 \mathbf{I})$$

and $\mathbf{m}_k$ variational parameter, $\eta$ given in advance is the mixing hyper-parameter, and small $\sigma^2$. We optimise over $\mathbf{m}_k$ the variational parameters of the random weight matrices; these correspond to the RNN's weight matrices in the standard view. The variational parameters $\mathbf{m}_k$ can be approximated using KL in Eq. (10) and $L_2$ regularisation [11].

Samples from the distribution $\hat{\boldsymbol{\theta}} \sim q(\boldsymbol{\theta})$ in evaluating the model corresponds to using dropout in the forward pass of the RNN by randomly zeroing elements in each weight matrix. The hyper-parameter $\eta$ is then the dropout ratio in the neural network.

To do final predictions by propagating the mean of each layer to the next corresponds to the posterior with $q(\boldsymbol{\theta})$. The prediction for unseen instances, then we can use Monte Carlo estimate, i.e.

$$p(r_{w^{(t+1)}}|w^{(t+1)}, \mathcal{R}_t, D_t, I)$$
$$= \int p(r_{w^{(t+1)}}|w^{(t+1)}, \mathcal{R}_t, D_t, I, f_{\hat{\boldsymbol{\theta}}}) q(\hat{\boldsymbol{\theta}}) \mathrm{d}\hat{\boldsymbol{\theta}}$$
$$\approx \frac{1}{N} \sum_{i=1}^{N} p(r_{w^{(t+1)}}|w^{(t+1)}, \mathcal{R}_t, D_t, I, f_{\hat{\boldsymbol{\theta}}}^{(i)}), \ f_{\hat{\boldsymbol{\theta}}}^{(i)} \sim q(\hat{\boldsymbol{\theta}})$$

where $\hat{\boldsymbol{\theta}}_k \sim q(\boldsymbol{\theta})$, that is, each element is drawn from a dropout operation that gives rise to the different function $f_{\hat{\boldsymbol{\theta}}}^{(i)}$. In the paper, for simplicity and brevity of presentation we use the samples from the true posterior, in Eq. 2.

**Additional Samples from the GuessWhat dataset**

In Table (4) we provide additional samples from the GuessWhat game played using our approach. It is interesting to note in these examples that our approach is able to find the objects that might be odd to see given the context. For instance, in the first row, there is a boat hanging from the wall. Given the past dialogue history and the image, finding a boat in the bedroom is unlikely. However, since our approach takes risky decisions for the choice of words in favour of extracting more information, it is able to find the boat hanging on the wall. In the subsequent two image samples, our approach is able to find the smaller objects utilising its dialogue with the oracle.

| Question-Answers | Image | Correct Answer |
|---|---|---|
| is it furniture ?   no<br>is it electronic ?   no<br>is it sink ?   no<br>is it plant ?   no<br>is it in right ?   yes | 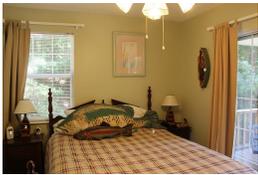 | boat (on the wall) |
| is it furniture ?   no<br>is it electronic ?   no<br>is it in left ?   no<br>is it in middle ?   yes<br>is it in front ?   yes | 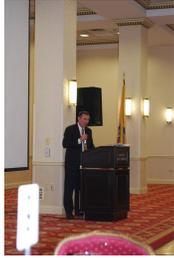 | person |
| is it person ?   no<br>racket?   no<br>ball ?   yes<br>is it in right?   no<br>is it in left?   no | 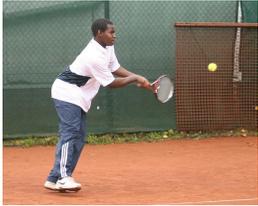 | ball |

Table 1. Samples of the dialogue from the GuessWhat dataset.



\end{document}